\documentclass[letterpaper, 10 pt, conference]{ieeeconf}  

\IEEEoverridecommandlockouts                              

\usepackage{formatting} 

\title{\LARGE \bf
DenseTact 2.0: Optical Tactile Sensor for Shape and Force Reconstruction
}

\author{Won Kyung Do, Bianca Jurewicz, and Monroe Kennedy III 
\thanks{Authors are members of the ARMLab in the Mechanical Engineering Department, Stanford University, Stanford, CA 94305, USA. 
{\tt\small \{wkdo, biancalj, monroek\}@stanford.edu.} 
The first author is supported by a fellowship from the Kwanjeong Educational Foundation. \edits{This work is supported by the National Science Foundation under Grant 2142773.} Youtube link for DenseTact 2.0: \url{https://youtu.be/5S74w0iSPz8}}
}

\begin{document}

\maketitle
\thispagestyle{empty}
\pagestyle{empty}

\begin{abstract}

Collaborative robots stand to have an immense impact on both human welfare in domestic service applications and industrial superiority in advanced manufacturing with dexterous assembly. The outstanding challenge is providing robotic fingertips with a physical design that makes them adept at performing dexterous tasks that require high-resolution, calibrated shape reconstruction and force sensing. In this work, we present DenseTact 2.0, an optical-tactile sensor capable of visualizing the deformed surface of a soft fingertip and using that image in a neural network to perform both calibrated shape reconstruction and 6-axis wrench estimation. We demonstrate the sensor accuracy of 0.3633$mm$ per pixel for shape reconstruction, 0.410$N$ for forces, 0.387$N\cdot mm$ for torques, and the ability to calibrate new fingers through transfer learning, which achieves comparable performance with only 12\% of the non-transfer learning dataset size.

\end{abstract}


\section{INTRODUCTION}

Robots must be able to manipulate objects with dexterity comparable to human performance in order to be effective collaborators in environments designed for humans. This requires both the physical design of a robotic fingertip that can accommodate complex objects as well as the modeling of the contact region relating deformation to calibrated shape and force measurements. Many robotic fingertip designs exist with various strengths and weaknesses with representative examples that include piezoelectric \cite{wettels2009multi}, optical \cite{Dong2017,doi:10.1177/02783649211027233, Hogan2018}, resistance \cite{cheng2009flexible}, capacity \cite{Huh2020} and hall effect \cite{yan2021soft}. Robotic fingertips can be broadly categorized in terms of transduction based sensors, where an electrical signal is caused by deformation and used to provide information of shape and force, versus optical-based sensors, where an image of the fingertip is observed and the deformation of the soft fingertip is correlated to shape and forces. For all of these sensors, the objectives are high contact resolution and calibration for both shape and forces. The challenge is obtaining high-resolution calibration for the shape and forces. Previous work explored high-resolution shape calibration for a vision based sensor \cite{9811966}, but a primary limitation was the inability to sense forces as well. A comparison of the proposed sensor and competitive sensor models is presented in Table \ref{table:comparison} with comparison metrics for sensor resolution, design shape, and force sensing modality.
\begin{figure}[t!]
	\centering
	\includegraphics[width=3.0in]{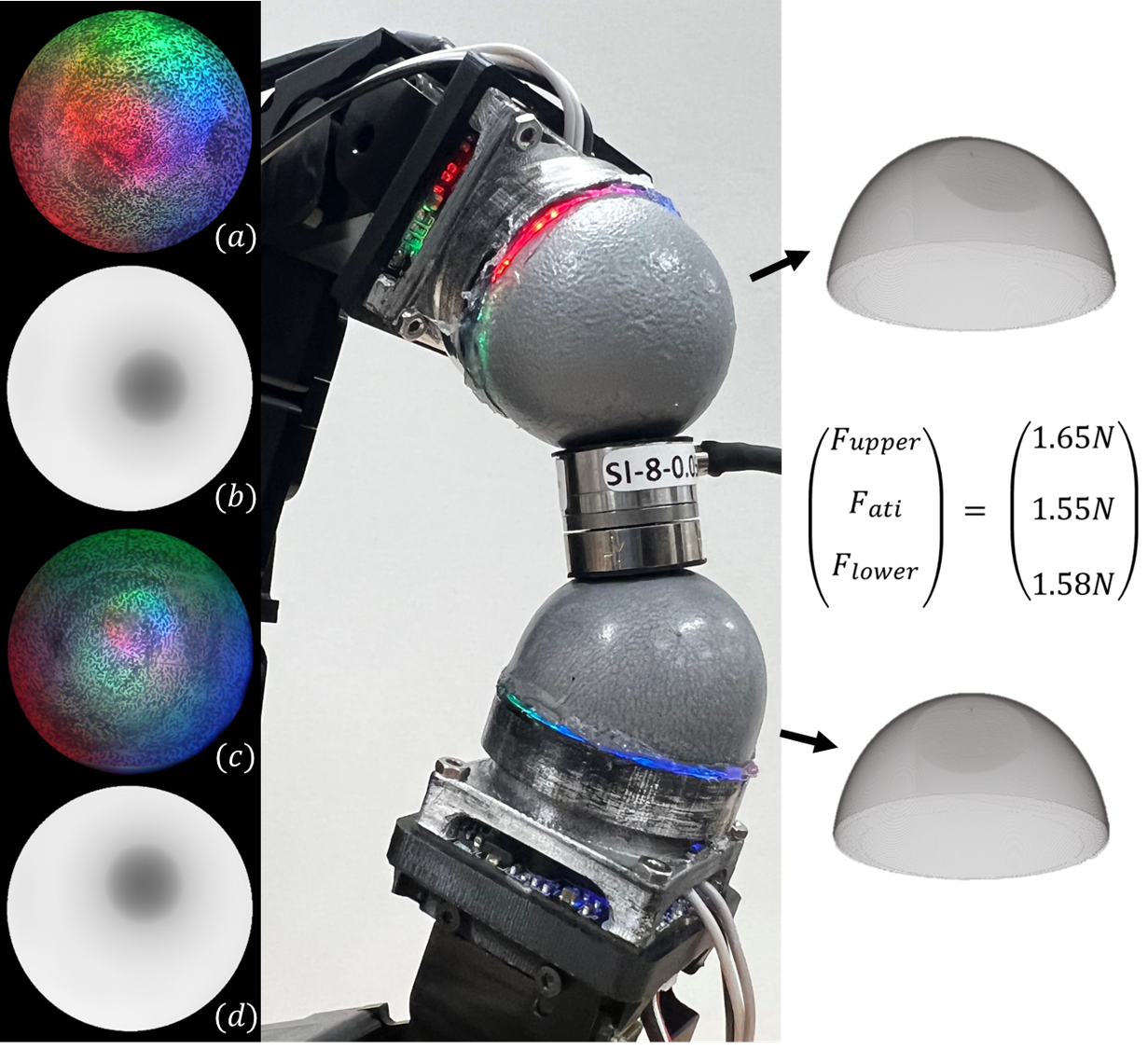}
  \caption{\textbf{DenseTact 2.0.} Sensors are pinching ATI Nano sensor\texttrademark. Image (a) and (c) shows the image taken from upper and lower sensors. Image (b) and (d) shows the corresponding resultant depth images. Right pointcloud represents the 3D reconstructed surface of sensor. Right middle matrix shows the estimated force from \edits{DenseTact 2.0s} and ATI sensor. }
  	\label{fig:main}
      \vspace{-0.71cm}
\end{figure}

To obtain the shape from the interior of an optical tactile sensor, the incidence angle of the interior projected light with the surface of the sensor must be used to approximate the normal to the surface, and with these normal's the sensor shape can be constructed from Poisson integration either directly or through an approximation method that leverages neural networks \cite{9811966}. To obtain forces, some studies leverage an inverse FEM model \cite{Ma2019}, use of an expressing marker, or leverage a skeletal structure for precise estimation via image input \cite{sun_soft_2022}. For all of these approaches, the deformation of the sensor must be tracked to correlate to observed forces. If a soft fingertip is used without skeletal structure \cite{taylor2021gelslim3, sun_soft_2022}, then the force distribution approximation may require a huge computational load due to the nature of the hyper-elastic material of the sensor. This can be mitigated with data-driven models, which can approximate forces from a single deflected image. To track the interior of the sensor, markers must adorn the interior in order to observe normal, shear and torsional deflection. However simple patterns such as dots can suffer the effects of aliasing under large deflection \cite{du2021high}. To accommodate this, we propose the use of a randomized, continuous pattern deposited on the surface of the sensor for tracking large deflection without aliasing.

Our contributions are as follows: 1) We present the physical design of \edits{the} DenseTact 2.0 which has upgrades in modularity, lighting design, and is approximately 60\% the size of DenseTact 1.0 \cite{9811966} with a novel surface patterning deposition technique.  2) We present the combination of a calibrated high-resolution shape reconstruction with a calibrated 6-axis wrench estimation. 3) We provide a comparison study between leading monocular depth estimation models with our model applied to both shape reconstruction and force estimation. 4) We show the effectiveness of transfer learning of our model for faster and more efficient training of future sensors with consistent geometry for calibration and deployment.

The paper is organized as follows: Sec. \ref{sec:sensor_design} presents the sensor design of DenseTact 2.0, Sec. \ref{sec:shape_force_data} presents the data preparation for force and shape estimation, \ref{sec:shape_force_estimation} presents the method of estimating force and shape with modeling, Sec. \ref{sec:results} shows results for the shape reconstruction model and force estimation model, and the conclusion and future work is discussed in Sec. \ref{sec:conclusion}.

\begin{table}[t]\centering
    \vspace{0.2cm}

\begin{tabular}{| c || c | c | c | c|}
 \hline
Name&Resolution&Force range&Shape\\ \hline
 \begin{tabular}{@{}c@{}}Gelslim 3.0 \cite{taylor2021gelslim3} \end{tabular}  & $640 \times 480$ & unspecified & \edits{full} \\\hline
  TaTa\cite{li2022tata} & $1280 \times 720$   & $\times$ & full \\\hline
  Skin sensor\cite{yan2021soft} & 0.1 mm   & $0\sim 3N, \,1$ & partial \\\hline
 Softbubble\cite{Alspach2019} & $224 \times 171$ & $\times$  &  full \\\hline
 Omnitact\cite{Padmanabha2020}& $400 \times 400$    & $\times$ & partial \\ \hline
 NeuTouch\cite{Taunyazov2020}& $39$&  $\times$  &  $\times$ \\ \hline
 Romero\cite{romero_soft_2020}&  $640 \times 480$  &  $\times$  & full \\ \hline
 Optofiber-sensor\cite{Baimukashev2020}&$61$ fibers & $0.03\sim 8N,\, 5$ & $\times$   \\ \hline
 GelTip \cite{gomes2020geltip} &  $\times$  & unspecified&  partial\\ \hline
 Digit\cite{lambeta2020digit}& $640 \times 480$  &  $\times$ & partial\\ \hline
 Insight \cite{sun_soft_2022} &   $1640$ $\times$ $1232$&$0.03\sim 2N, \,5$ & partial \\ \hline
 DenseTact 1.0 \cite{9811966} &   $800$ $\times$ $600$ & $\times$ & full \\ \hline
 \hline
 \textbf{DenseTact 2.0} & \textbf{$1024$ $\times$ $768$} & $-11 \sim 3N,\, 6$ & full \\
 \hline

\end{tabular}
\caption{\textbf{Related Work.} Table shows resolution of sensor, sensing range and dimension of the force, and availability of shape reconstruction. `Partial' means the sensor does not estimate the depth of the entire sensing area, or estimates only the position of contact.}
\label{table:comparison}
     \vspace{-0.7cm}

\end{table}

\section{DenseTact 2.0 Sensor Design}
\label{sec:sensor_design}
\subsection{Design Criteria}

A tactile sensor with a highly-deformable gel has a clear advantage for the vision-based approach. Gel deformation not only enables collecting the information of the contact object, but also easily tracks features, even with small indentation. In order to extract as much geometrical and force information from the single image, the sensor requires more attractive features. Furthermore, the in-hand manipulation is more prone to happen with a compact sensor size. To deal with these issues, we augmented the design of DenseTact \cite{9811966} with following features: 1) Reduced sensor size while maintaining highly-curved 3D shape. 2) Modular design \edits{using} off-the-shelf materials for easy assembly and resource-efficiency. 3) Enriched features with randomized pattern on the surface for force estimation.

  \begin{figure}[t!]
  
    \vspace{0.2cm}
      \centering
      \includegraphics[width = 3.3 in]{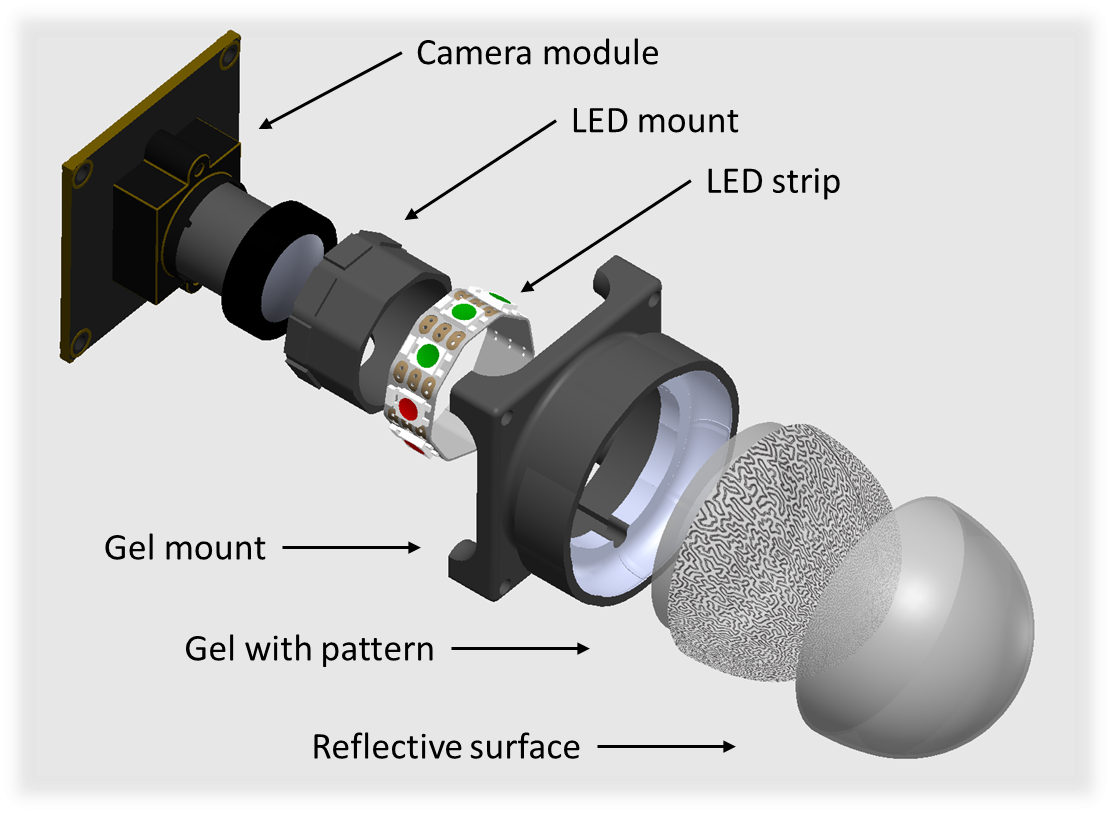}
      \caption{ \textbf{DenseTact Design}. Exploded view of DenseTact 2.0. The gel mount reflects the light from LED while the gel is covered with the pattern and reflective surface. }
      \label{fig:CAD}
      \vspace{-0.8cm}

  \end{figure}

\subsection {Gel Fabrication with Randomized Pattern } \label{sec:fab}

The fabrication process of the gel \edits{follows} three steps: 1. making a gel base \cite{9811966}, 2. printing a randomized pattern on the surface of the gel, and 3. covering the gel with a reflective surface. 

\subsubsection{Gel Base}
The material of the gel is the same material (49.7 shore OO hardness) and similar mold shape as in \cite{9811966}, while the compact hemispherical shape has a 31.5mm \edits{diameter}. Compared to the DenseTact 1.0, we increased the contact area between the gel mount and lens to be more durable - the contact area - volume ratio of the DenseTact 1.0 is $0.0707 mm^{-1}$ ($Area$ / $Vol = 3,264.3 mm^2$ / $46,173 mm^3$), and the ratio of DenseTact 2.0 is $0.1229 mm^{-1}$ ($Area$ / $Vol = 1,443.4 mm^2$ / $11,746 mm^3$).

\subsubsection{Randomized Pattern on the Gel Surface}
The randomized pattern can hold more information for extracting features from a single image, such as continuous deformation output or non-aliasing problem. Marker-based approaches seen in most tactile sensors are hard to deal with the aliasing problem with large deformation. One other approach such as the use of a randomly colored pattern \cite{zhang2021multidimensional} enables intrinsic features to follow, but it is only applicable for sensors with planar surfaces, and the RGB channel can interfere with the pattern itself. Furthermore, the pattern requires to have a unique pattern to avoid the aliasing problem and maintain a balanced density between the pattern and background to extract the feature from the surface deformation. 

To create the unique pattern, we first distribute points on the 2D planar surface using the voronoi stippling technique \cite{rougier2017re} and randomly connect all points. Connecting the array of points can be considered as the Traveling Salesman Problem (TSP), a classic algorithm for finding the shortest route to connect a finite set of points with known positions. We connect all points with a TSP solver \cite{applegate2006concorde}, convert the solution as an image file, and extract the unique pattern using 8,192 points on a $25mm \times 25mm$ size square. 

We printed a stamp plate of the randomized pattern using a laser cutter with a depth of 0.03mm. Next, we spread an ink on the plate, where the ink is composed of a silicone base with black ink (Smooth-on Psycho Paint\texttrademark{} and pigment, the ratio of silicone base to ink is 5:1). Then we scrape the ink on the plate so that the ink only remains on the \edits{imprinted} part of the stamp. Next, we press the cured gel onto the ink and distribute the pattern evenly by contacting all parts of the surface only once. The result of the printed pattern is shown in input images at Fig. \ref{fig:main}.

\subsubsection{Reflective Surface} The reflective surface is made from a mixture of silicone paint and silver silicone pigment with the ratio of 2:1. 0.5$\%$ of the solution of Thivex thickening solution is added to the mixture. Then, the mixture is placed in a vacuum chamber to remove any air bubbles that may be present from mixing the materials together. The method of applying the reflective surface improved from versions DenseTact 1 to 2.0 as the application time reduced from 2 hours to 30 minutes. This is done by leveraging a paint dipping technique as opposed to airbrushing. To execute this method, a suction cup is used to grip the gel, which is then dipped into the silicone ink mixture. The gel is dipped in the ink a total of three times and a heat gun is used to cure the paint after each dip. With this method, users can easily \edits{repair} from possible abrasion created through extended gel usage by dipping the gel into the ink solution whenever necessary.

\subsection{Sensor Fabrication} 

The bottom part of the sensor contains a camera, LED mount, LED strip, and a gel mount covered with mirror-coating. The sensor's exploded view is shown in Fig. \ref{fig:CAD}. 

\subsubsection{Illumination with Mirror-Coated Wall}

The major requirement for a vision-based sensor is illumination. Because of the compact size of the sensor and the LED being a point light source with a limited angle of light emission, the LED strip with a single RGB channel LED had a limitation when the sensor became smaller. Therefore, we implemented the new illumination system with a mirror-coated wall while still maintaining the simple assembly feature.

Instead of using 3 LED lights as in \cite{9811966} or other tactile sensors, we utilized 9 LED lights (3 LEDs for each color- red, green, and blue) from an LED strip (Adafruit Mini Skinny NeoPixel \texttrademark{}) while controlling the intensity of each LED. As shown in Fig. \ref{fig:CAD}, the LED surrounds the camera while facing outside. An equivalent distance between each LED with more lights allows the sensor to get an equal distribution of lights. Furthermore, the increased brightness makes the sensor more resistant to external lights. 

The 3D-printed gel mount reflects the lights to the gel through the mirror-coated surface. To develop the mirror-like effect on the side, we flattened the surface of the gel mount with XTC-3D\texttrademark{} and coated it with the mirror-coating spray. Finally, the lights on the LED pass through the opposite side of the gel (see the input image in Fig. \ref{fig:main}). 

\subsubsection{Sensor Assembly}

We modularized the sensor into three parts - gel with gel mount and lens, LED module, and camera module. Each module is easily replaceable while the other modules remain intact. The gel, gel mount and lens are firmly attached through sil-poxy adhesive\texttrademark{} and Loctite Powergrab\texttrademark{} Crystal Clear adhesive. The gel module and LED module is fixed through the 4 screws with camera module. The user can simply unscrew and replace either the camera, LED, or gel module. Since the sensor has more contact area-volume ratio, the durability increased even with the modularized design.

We chose to use the camera module Sony IMX179 (30fps) and M12 size lens with the field of view 185 $\deg$ degree for easy replacement. \edits{The distance of focal length of the camera is manually set to optimize the focal length for the expected deformation. }The final size of the sensor including the camera is $W \times D \times H = 32 \times 32 \times 43 mm$ with the weight 34g. The cost of the sensor became cheaper because of the smaller LED strip (\$3.75), gel part (\$3), and camera\edits{ \ LED} mount (\$1.5) with the same price of the camera system (\$70). 

  \begin{figure}[t!]
      \vspace{0.2cm}

      \centering
      \includegraphics[width = 3.0in] {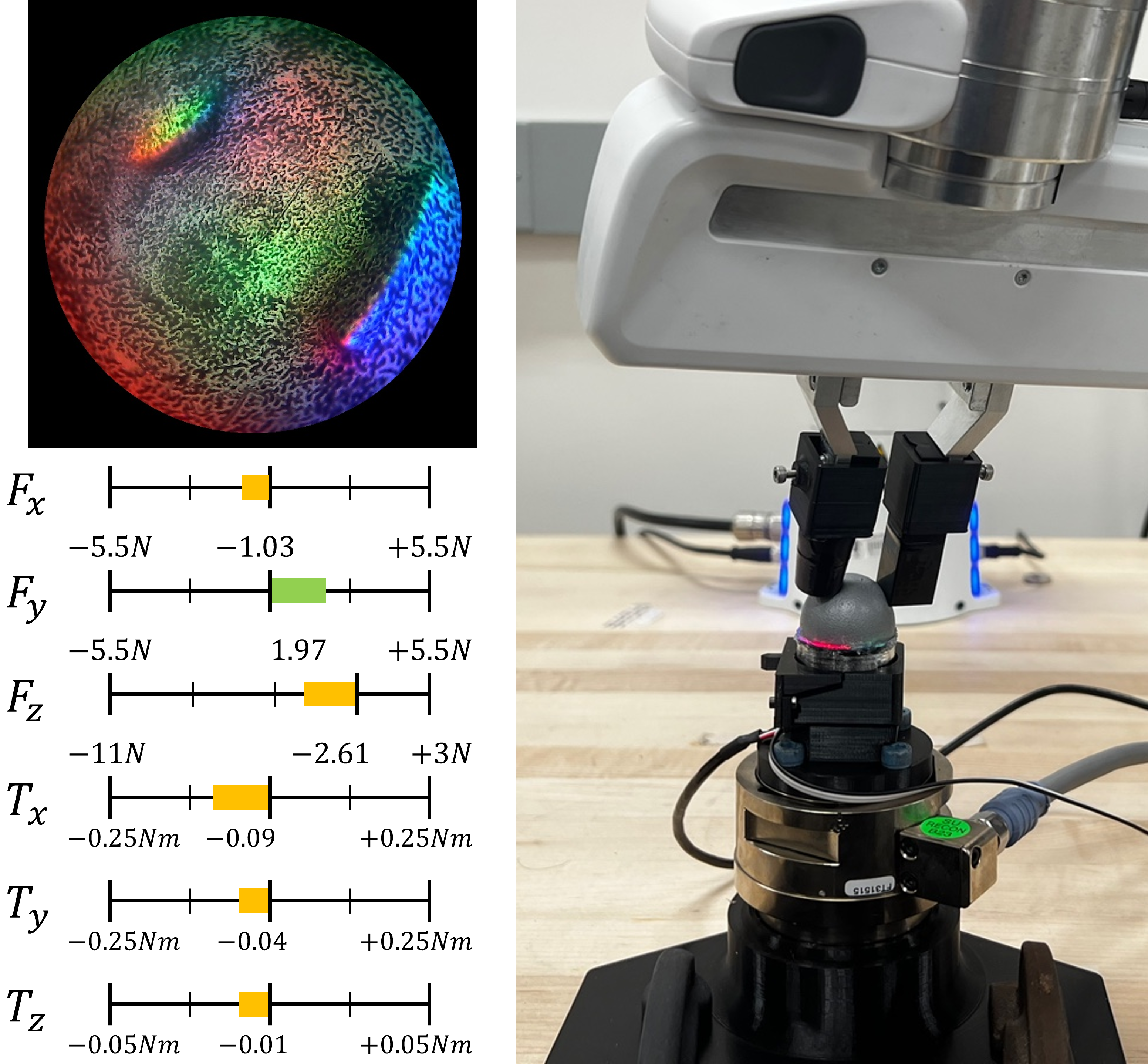}
      \caption{\textbf{Force Data Collection.} Densetact 2.0 has been pushed under an ATI Gamma sensor with the Franka\texttrademark{} arm, where the grippers are covered with indenters.}
      \label{fig:datacollect}
          \vspace{-0.7cm}

  \end{figure}

 \begin{figure*}[t]
      \centering
          \vspace{0.2cm}

      \includegraphics[width = 6.1 in]{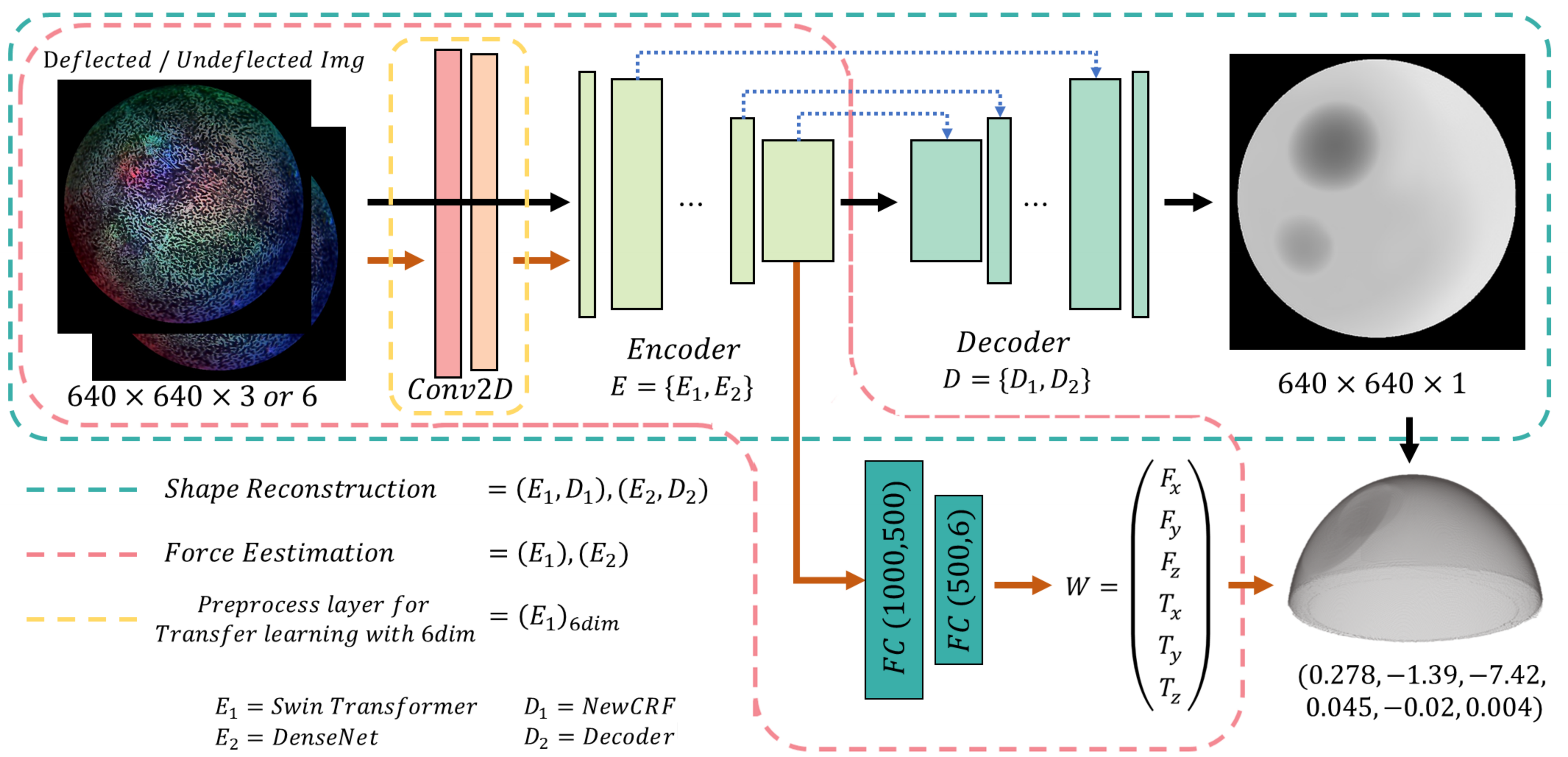}
      \caption{\textbf{DenseTact 2.0 Algorithm}. The sensor interior is the input to the encoder-decoder network for shape estimation model (blue box). Force estimation models (red box) consist of encoder and fully-connected layers. Transfer learning model with 6 dimension (includes yellow box) takes deflected and undeflected image as an input. (Model references: $E_1$-\cite{liu2021swin}, $E_2$-\cite{huang2017densely}, $D_1$-\cite{yuan2022neural}, $D_2$ -\cite{alhashim2018high})}
      \label{fig:process}
          \vspace{-0.60cm}

  \end{figure*}

\section{Data preparation for force and shape estimation}\label{sec:shape_force_data}

\subsection{Data Collection Process for Shape Reconstruction}

The dataset for shape estimation has been collected in a similar manner as the DenseTact 1.0 with more autonomy \cite{9811966}. While utilizing the CNC machine with a stepper motor for precise movement, we implemented an encoder for the stepper motor and a limit switch for an autonomous procedure. The sensor is attached to the stepper motor side with a mount. The mount ensures the center of the sensor is aligned with the rotational center of the stepper motor. 

In order to collect more datasets in one process, 21 indenters, a Stl model which covers the entire sensor surface, are placed in a $3 \times 7 $ grid on the plate of the CNC machine. Each row contains the same shape of indenters, each with a different orientation along a random axis. The rotation axis is aligned with the center of each indenter and placed in the xy plane. Therefore, each row shows different orientations by rotation on the x and y axes, while the z axis rotational difference is provided from the stepper motor. As a result, each data collection procedure generates up to 8,400 image datapoints (21 indenters $\times$ 400 steps\slash rev) without human intervention. 

Consideration of the bulging effect is a major improvement in the data collection process. Since the silicone gel is an  incompressible, and hyper-elastic material, deformation causes bulging on regions not in contact with the indentor. For shape reconstruction, we account for this in the generated contact shape Stl file. In this way, the sensor is exposed to a more natural deformation.
The size of captured image from camera has been increased into ($1024 \times 768 \times 3$), which leads the final image size into ($640 \times 640 \times 3$). After collecting the input image, the depth is reprocessed from the corresponding Stl model through a Gaussian Process with a ray casting algorithm \cite{9811966}. The Stl files are available on github \footnote{\url{https://github.com/armlabstanford/DenseTact}}.

\edits{The default distance of the sensor in the dataset is 15.5mm, and its radial distance ranges from 12.23mm to 16.88mm, while the opposite side of the sensor bulges when deformed, resulting in a larger depth value than the hemisphere radius.}
Allowing for a margin around 0.05mm, we normalized the depth value from 12.23mm to 16.88mm (4.64mm range) into 0-255 pixel values for ground truth depth images. Finally, the 1 pixel increment corresponds to a 0.0182mm increment in depth value. We collected a dataset for two sensors - the sensor 1 has 38,909 training and 1,000 test configurations, and the sensor 2 has 20,792 training and 1,000 test configurations. Test configurations for each sensor are recorded with an unseen indenter from the training dataset. The datasets have a total 8.7GB and 6.8GB size for the sensor 1 and 2.

\subsection{Dataset Collection for Force Estimation}
The force dataset has been generated through randomly pushing sensors with the Franka arm. This method allowed us to collect the dataset with no constraints on the pose of the franka arm. The right image in Fig. \ref{fig:datacollect} shows the configuration of the force dataset collection where the DenseTact 2.0 is mounted on the ATI Gamma sensor (SI-65-5). 
We created 10 different objects to push the sensor and collected the dataset while either attaching an object on each gripper finger or by gripping an object. The set of objects includes cylindrical shapes, spherical shapes and daily objects such as nuts. All joint positions including the position of the gripper fingers were recorded during dataset generation in the rate of 1,000 paths per second. The recorded motion of the Franka arm for calibration makes multiple sensor calibration easier by requiring less human intervention. 

During online dataset collection, we \edits{filter out duplicated sequential images which do not show significant change using Peak signal-to-noise ratio (PSNR) as a similarity metric}. The ring buffer collects the image up to 5 current images and applies the following threshold - $ PSNR(Img_{curr}, Img_{prev,i}) < 0.9 , \text{ where }  i = 1,\,...\, ,5 $. The dataset has been collected within the range specified in the left part of the Fig. \ref{fig:datacollect}, 
where the unit of force and torque are $N$ and $N\cdot m$. the left image and force distribution in Fig. \ref{fig:datacollect} shows the collected input image and corresponding force and torque data. The final dataset has been normalized between each force and torque range. The dataset has been collected for two sensors - the sensor 3 has 38,909 training force points and 1,000 test points, and the sensor 4 has 20,792 training points and 1,000 test points. Test points are collected with different shape as the pushing objects used in each training points. Total size of each dataset is 7.5GB and 2.7GB, respectively.

\section{Algorithms for shape and force reconstruction}\label{sec:shape_force_estimation}
\subsection {Algorithms for Shape Reconstruction}
While the randomized pattern on the surface adds more features for continuously tracking surface movement, reconstructing the sensor surface requires learning features such as the deflected part's location, or surface normal based on the LED position. The position of the random pattern also gives the dynamic movement of the sensor, which requires the networks to learn more features from a single image. Therefore, we compared two network models to reconstruct the shape of the sensor surface.

\subsubsection{Network with Swin Transformer and NeWCRF}
The Vision Transformer (ViT) is a well-known model from transformer-based architecture for image classification \cite{dosovitskiy2020image}. While ViT splits an image into patches and train position embedding for each image patch, the Swin transformer builds the feature maps hierarchically with lower computational complexity because of a localized self-attention layer \cite{liu2021swin}. Our input image contains closely-related information relation between neighbor pixels. Therefore, the path embedding with hierarchical feature maps between each layer can better connect information between the indented and opposite parts. 

Once the input image has been trained with swin transformer as the encoder part, the decoder is also important for correlated embeddings. Recently, models using a classification model to boost the performance of depth estimation, such as binsformer\cite{li2022binsformer} or adpative bins \cite{bhat2021adabins}, perform well with the monocular depth estimation. However, Neural window FC-CRFs (NeWCRF) reaches the same performance by applying Conditional Random Field (CRF) on the decoder part to regress the depth map by utilizing fully-connected CRFs on each split image part (window) \cite{yuan2022neural}. Therefore, we chose the Swin Transformer with NeWCRF decoder among the state of the art models for monocular depth estimation. 

As shown in the green part of Fig. \ref{fig:process}, the network gets input as $640 \times \edits{640} \times 3$. Without using the pretrained model, we normalized both the input and ground truth depth images from 0 to 1 (ground truth originally 0-255). We utilized 4 swin-transformer blocks on the encoder part using window size 20, the number of patches from one image. The predicted model is compared with the ground truth using the Scale-Invariant Logarithmic loss (SILog loss) with 0.85 as variance minimizing factor \cite{eigen2014depth}. The model is trained for 21 epochs with the batch size of 8 while the learning rate starts from $2 \times 10^{-5}$ on the 4 Nvidia A4000 GPUs. The model took about 36 hours for training.



\subsubsection{Densetact Net Position}

The above model is compared with the Network from \cite{9811966} without resizing the image. As shown in Fig. \ref{fig:process}, the network consists of an encoder as Densenet \cite{huang2017densely}, and a simple decoder with skipped connections \cite{alhashim2018high}. The final result has been upsampled by the upsampling layer to get the $640 \times 640 \times 1$ as an output depth image. Unlike the above model, the input and ground truth are un-normalized. The network is trained without any prior or pretrained model and used the reciprocal of the depth for structural similarity loss. 

By comparing the above model with ours, we can show 1) whether the random pattern blocks the estimation result and 2) how many model parameters are enough to estimate the depth or force estimation. The training runs for 25 epochs with batch size 8. The learning rate is set to $1\times 10^{-4}$, where the model took about 16 hours for training. 

\subsection{Algorithm for Force Estimation}

The network model for force estimation utilized each encoder part of the above two models. The network structure for force estimation is illustrated in Fig. \ref{fig:process}. After passing either the Swin transformer encoder or the Densenet-based encoder, two fully-connected layers shrink the channel size from 1,000 to 500, and from 500 to 6, which corresponds to the 6 wrench inputs. The learning rate starts decreasing from $2 \times 10^{-5}$. Both models are trained with batch size 8 for 22 epochs. The training has been done for each sensor dataset. 

  \begin{figure}[t!]
      \centering
          \vspace{0.1cm}

      \includegraphics[width = 3.1 in ] {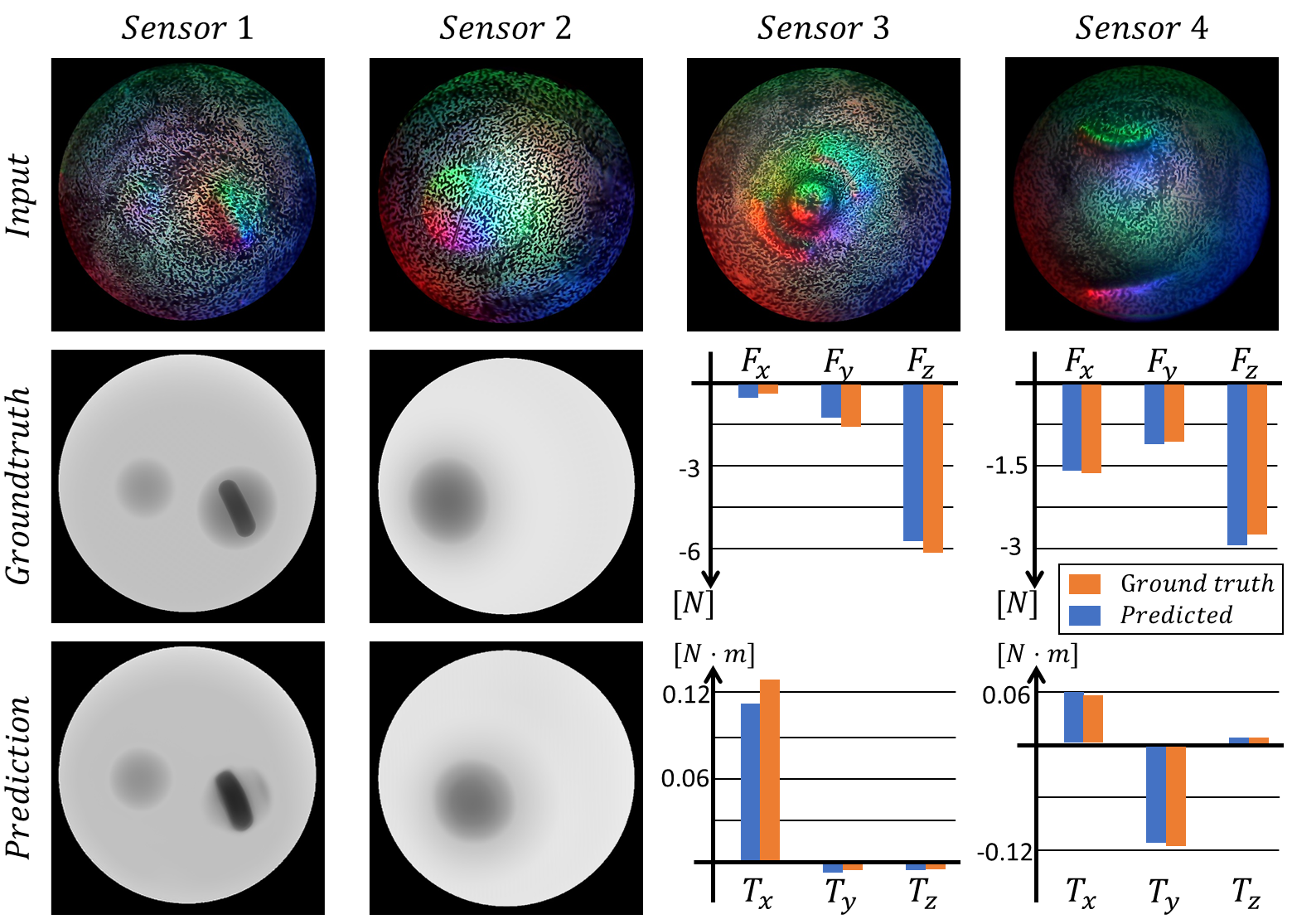}
      \caption{ \textbf{Shape and Force Reconstruction Performance.} Examples from test set for sensor shape and force reconstruction.}
      \label{fig:dataresult}
          \vspace{-0.6cm}

  \end{figure}

\subsection{Data-efficient Training with Transfer Learning }

One major disadvantage of the data-driven approach in tactile sensing estimation is that the data collection process is much longer than the modeling-based approach. However, the pre-trained model from multiple sensor datasets can significantly reduce the burden for the calibration process. We proposed a two network structure for transfer-learning model for our sensor. 

The first model, 3-dim model, has a network identical to the Densenet-based encoder, which takes input as $640 \times 640 \times 3$. The training datasets are combined for both sensors, while the test sets are also combined from each sensor's test dataset. Finally, the total number of configuration training data points is 59,701 and for the test dataset is 1,200, where 600 data points are extracted from each sensor. 

The second model, 6-dim model, simultaneously takes the undeflected image and deflected image of the sensor so that the input becomes $640 \times 640 \times 6$. The blue part in Fig. \ref{fig:process} indicates the layer applied for the model with 6-dimensional inputs. Two blocks of convolutional, batchnorm, and relu layers reduces the number of channels from 6 to 4, and from 4 to 3. Finally, the input has been passed to the same Densenet-based encoder. Both models are trained for 30 epochs with a batch size of 16.
After getting the pretrained model, we compared the loss of each model by using a small dataset of the another, \textit{unseen} sensor input for both force and position dataset.
The size of the small dataset is 4,643 and the dataset is pushed with a single object. 

  \begin{figure}[t!]
      \centering
          \vspace{0.2cm}
      \includegraphics[width = 3.3 in ] {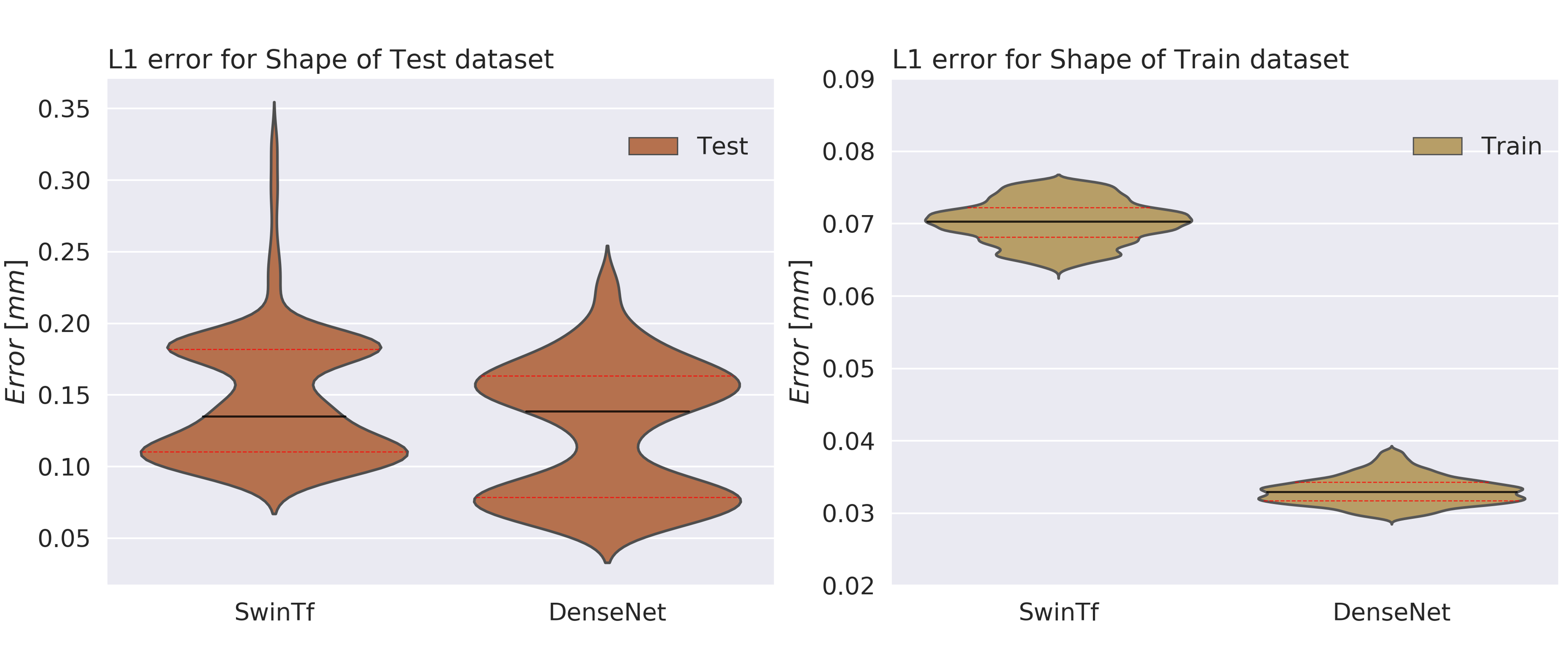}

      \caption{ \textbf{Violin plot of shape reconstruction model.} Left and right image compares L1 error for model SwinTF and DenseNet on test and train dataset, respectively. }
      \label{fig:violin_pos}
          \vspace{-0.6cm}

  \end{figure}
  
    \begin{figure}[t!]
      \centering
          \vspace{0.2cm}

      \includegraphics[width = 3.3 in ] {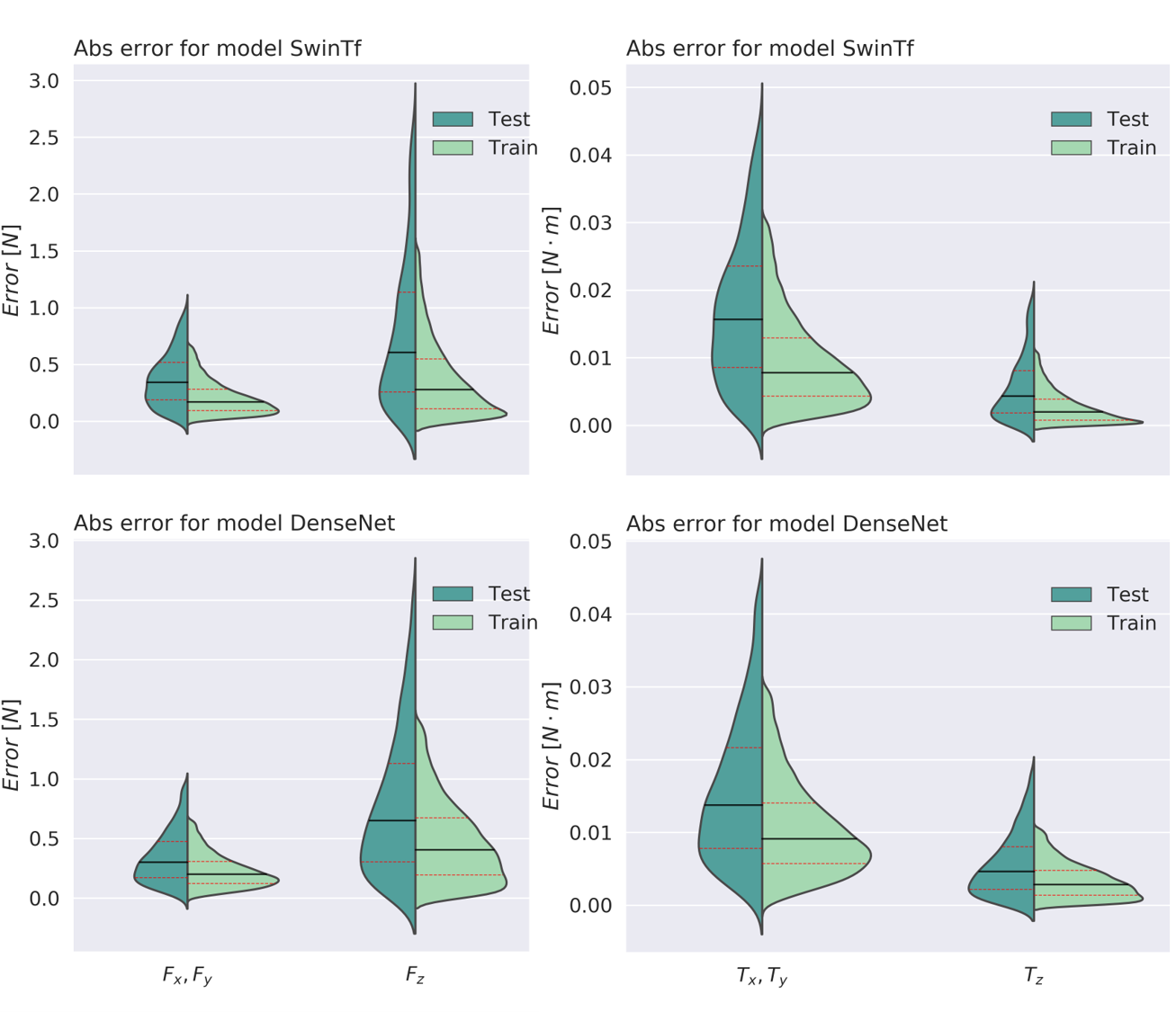}
      \caption{ \textbf{Violin plot of force estimation model.} Upper and lower images show the absolute error of train and test dataset from model SwinTF and DenseNet. Left and right side shows the plot of $F_{x,y}, F_z$, $T_{x,y}, T_z$, respectively.}
      \label{fig:violin_force}
          \vspace{-0.7cm}

  \end{figure}
  
\section{Results and Discussion}
\label{sec:results}
\subsection{Validation of Models}

Fig. \ref{fig:violin_pos} and Fig. \ref{fig:violin_force} show the network evaluation with violin plots for position and force for all training and test datasets, respectively. The qualitative performance of the model is shown in Fig. \ref{fig:dataresult}. The mean of the L2 loss for SwinTF model, a shape model with encoder as Swin transformer and decoder as NeWCRF, with shape test set is $(sen_1 , sen_2) = (0.370mm, \,0.584mm), \, err_{tot} = 0.4769mm$, and the L2 loss for the DenseNet model, a shape model with encoder as DenseNet and decoder as skipped decoder, is $(sen_1 , sen_2) = (0.282mm, \,0.445mm), \, err_{tot} = 0.3633mm$. Both results show that the DenseNet model outperforms the SwinTF model. 

The force model also shows that the DenseNet model, a force model with encoder as DenseNet, performs slightly better than the SwinTF model, a model with encoder as Swin transformer. The total absolute mean error of force of each sensor on the test set for the SwinTF model is $(sen_3 , sen_4) = (0.426N, \,0.436N)$, and force error for the DenseNet model is $(sen_3 , sen_4) = (0.409N,\, 0.410N)$. For the torque, the absolute mean errors are $(sen_3 , sen_4) = (0.416 \,N\cdot mm,\, 0.438\,N\cdot mm)$ for the SwinTF model, and $(sen_3 , sen_4) = (0.395 \,N\cdot mm,\, 0.377\,N\cdot mm)$ for the DenseNet model.
In conclusion, DenseTact 2.0 performs the shape reconstruction with an absolute mean error of 0.3633mm with the DenseNet model, and performs force estimation with an absolute mean error of $(e_{force}, e_{torque}) = ($0.410N , 0.387 $N\cdot mm)$.

The result of transfer learning indicates that the model that gets 3-dimensional input works similarly for both big datasets and small datasets. A 3-dim model trained with a small force dataset on top of a pretrained model has an absolute mean error of normalized wrench of 0.05163, whereas the 6-dim model with a pretrained model is 0.05143. However, the 3-dim model converged faster (converged at 7,200 steps) than the 6-dim model (converged at 8,400 steps). 3-dim model with a position dataset has converged faster (converged at 6,900 steps) with an rms error of 0.2275mm, than the 6-dim model (converged at 8,100 steps) with rms error of 0.2394mm. 

\subsection{Discussion}

The results above \edits{demonstrate} that the randomized pattern on DenseTact 2.0 performs well with both shape reconstruction and force estimation. The SwinTF model for shape reconstruction has about 270 million parameters, while the DenseNet model for shape has 44 million parameters. The forward path for the SwinTF model takes 0.124s per step, while the DenseNet model takes 0.04s per step on Nvidia 3090 GPU. Therefore, the Densetact input requires fewer parameters for training and works better with the DenseNet model. The state of the art model for monocular depth estimation might perform well on the daily objects and general images, but the input image for DenseTact 2.0 requires solving the relation between each pixel by estimating the position of the LEDs and tracking the pattern deflection. 

Both transfer-learning models \edits{demonstrated} that the model works with a smaller dataset, which is about 10\% of the original training dataset size. The training time only took 1 hours and 1.8 hours for force and position, respectively, for 30 epochs with a batch size 10 on Nvidia A4000 GPU. Considering that the data calibration process can be done easily with the ATI force sensor, the calibration time is comparable with the model-based sensors.

\section{Conclusion}
\label{sec:conclusion}

This paper presents DenseTact 2.0, a compact-size calibrated shape and force sensor with an very soft gel, which is capable of reconstructing surface shape and force estimation with high resolution. The modularized design and compact size of DenseTact 2.0 enables versatile in-hand manipulation as well as easy assembly. We leverge a marker deposition algorithm which avoids aliasing under large sensor deformations. 
We benchmarked multiple models and show the benefit of using transfer learning which allows us to use 12\% of the original dataset size. 
Our future direction is estimating force distribution from the single image output while applying the transfer learning feature as well. 


  \bibliographystyle{./IEEEtran} 
  \bibliography{./IEEEexample}

\end{document}